\documentclass[11pt,a4paper]{article}
\usepackage[hyperref]{emnlp2020}
\usepackage{times}
\usepackage{latexsym}
\usepackage{amsmath}
\usepackage{multirow}
\usepackage{makecell}
\usepackage{array}
\usepackage{hhline}
\usepackage{booktabs}
\usepackage{algorithmic}
\usepackage{setspace}
\usepackage{graphicx}
\usepackage{stackengine}
\usepackage{xcolor}
\usepackage{float}
\usepackage{soul}
\usepackage{bm}
\usepackage{todonotes}
\usepackage{xurl}
\newcolumntype{L}[1]{>{\raggedright\let\newline\\\arraybackslash\hspace{0pt}}m{#1}}
\newcolumntype{C}[1]{>{\centering\let\newline\\\arraybackslash\hspace{0pt}}m{#1}}
\newcolumntype{R}[1]{>{\raggedleft\let\newline\\\arraybackslash\hspace{0pt}}m{#1}}

\usepackage{tabularx} 

\aclfinalcopy 

\title{Factual Error Correction for Abstractive Summarization Models}

\author{Meng Cao \qquad Yue Dong \qquad Jiapeng Wu \qquad Jackie Chi Kit Cheung \\ \\
    School of Computer Science, McGill University, Montreal, QC, Canada \\
    MILA, Montreal, QC, Canada \\ \\
    {\small \{\tt meng.cao@mail, yue.dong2@mail, jiapeng.wu@mail, jcheung@cs\}.mcgill.ca}}

\date{}

\begin{document}
\maketitle
\begin{abstract}
Neural abstractive summarization systems have achieved promising progress, thanks to the availability of large-scale datasets and models pre-trained with self-supervised methods. 
However, ensuring the factual consistency of the generated summaries for abstractive summarization systems is a challenge. We propose a post-editing corrector module to address this issue by identifying and correcting factual errors in generated summaries. The neural corrector model is pre-trained on artificial examples that are created by applying a series of heuristic transformations on reference summaries. These transformations are inspired by an error analysis of state-of-the-art summarization model outputs. Experimental results show that our model is able to correct factual errors in summaries generated by other neural summarization models and outperforms previous models on factual consistency evaluation on the CNN/DailyMail dataset. We also find that transferring from artificial error correction to downstream settings is still very challenging\footnote{Our data and code is available at \url{https://github.com/mcao610/Factual-Error-Correction}}.

\end{abstract}

\section{Introduction}
Self-supervised methods have achieved success in a wide range of NLP tasks, and automatic summarization is no exception \citep{liu-lapata-2019-text, lewis2019bart, zhang2019pegasus, shi-etal-2019-leafnats, fabbri-etal-2019-multi}. These state-of-the-art abstractive summarization models typically finetune pre-trained transformer-based models on a summarization dataset \cite{vaswani2017attention}. 
Despite significant improvements over previous methods in terms of automatic evaluation scores such as ROUGE \citep{lin-2004-rouge}, ensuring factual consistency of the generated summary with respect to the source remains challenging. For example, \citet{cao2018faithful} claims that about 30\% of summaries generated by abstractive models contain factual errors, which greatly limits their practicality. 

Different approaches have been proposed to detect or ensure the factual consistency of generated summaries, including using fact extraction or applying attention on fact triples \cite{cao2018faithful, zhang2019optimizing, goodrich2019assessing}, applying natural language inference or question answering models for consistency checking \cite{falke2019ranking, li2018ensure,  wang2020asking} and training the model on artificial datasets \cite{kryscinski2019evaluating}. Most of these approaches either require a high-quality fact extraction model or they only focus on factual consistency \emph{evaluation}. Improving factuality \emph{correction} by editing inconsistent parts in generated summaries is a direction that has not been explored much.

\begin{table}[t]
\small
\renewcommand{\arraystretch}{1.3}
\setlength\tabcolsep{2.5pt}
\centering
\begin{tabular}{|p{7.3cm}|}
  \hline
  {\bf Source}: \\
    Jerusalem (CNN)The flame of remembrance burns in Jerusalem, and a song of memory haunts Valerie Braham as it never has before. This year, Israel's Memorial Day commemoration is for bereaved family members such as Braham. ``Now I truly understand everyone who has lost a loved one,'' Braham said. (...) \\
  \hline
  \emph{Original}: {\bf France's } memorial day commemoration is for bereaved family members as braham. \emph{(inconsistent)} \\
  \emph{After Correction}: {\bf Israel's } memorial day commemoration is for bereaved family members as braham. \emph{(consistent)}
  \\
  \hline
\end{tabular}
\caption{\label{table:correction_example} An example of an inconsistent system-generated summary and the output summary from our correction model. In this case, ``France'' is successfully corrected as ``Israel''.}
\end{table}

In this work, we propose a model to improve the factual consistency of system summaries with \textit{post-editing correction} (Table \ref{table:correction_example}). Our model takes a draft summary that is generated by an abstractive summarization model and produces a corrected final summary, conditioned on the source document.  In addition, our trained corrector can be used as an evaluation model for factual consistency of abstractive summaries, with the assumption that a generated summary is inconsistent if our corrector decides to make edits. To teach the model to correct errors, we train it with artificial data that has factual errors introduced using heuristics proposed by \citet{kryscinski2019evaluating}. 


The empirical results based on automatic and human evaluations indicate that our model not only corrects factual errors in summaries, it is also a reliable factuality evaluation model.
In a downstream setting where we apply the corrector to the output of an abstractive summarizer, we find that our corrector is able to accurately correct errors in the generated summaries. However, the overall recall on correcting factual errors in real system summaries remains low, suggesting the errors introduced by heuristics have a different distribution than errors made by abstractive summarization systems. 




\section{Background and Related Work}
Previous work on factual consistency in abstractive summarization can be divided into two categories: abstractive summarization models tailored towards factual consistency \citep{cao2018faithful,zhang2019optimizing,li2018ensure}, and evaluation models for factual consistency in abstractive summarization \citep{goodrich2019assessing,falke2019ranking,kryscinski2019evaluating,wang2020asking}. 

\citet{cao2018faithful} proposed a dual attention module in an abstractive summarizer that attends to both the source document and to relation triples extracted from the document. \citet{zhang2019optimizing} propose to improve their abstractive summarization model by optimizing fact scores defined in radiology reports with reinforcement learning methods. \citet{li2018ensure} jointly train their model's encoder on summarization and NLI tasks. \citet{guo-etal-2018-soft}  train an abstractive summarization system with the auxiliary tasks of question and entailment generation and show that their generated summaries are less likely to produce extraneous facts. \citet{kumar-cheung-2019-understanding} show that neural abstractive summarizers often assign higher posterior likelihood to perturbed contrastive summaries that are inconsistent with the source text than to human-written gold-standard ones. Concurrently to our work, \citet{zhu2020boosting} recently proposed a fact-aware summarization model that uses a knowledge graph. They use a pre-trained corrector module to modify generated summaries. Concurrent to our work, \citet{dong-2020-multifact} proposes factual correction models that leverages knowledge learned from question answering models  via span selection. Their models employ single  or multi-masking strategies to either iteratively or auto-regressively replace entities.



In terms of evaluating abstractive summarization models for factual consistency,
\citet{goodrich2019assessing} proposed a metric to check factual consistency by checking the overlapped fact triples between a source document and generated text on Wikidata. \citet{falke2019ranking} shows that  factual error detection is a difficult task on its own and  adapting entailment models for factual error detection do not offer the desired performance. \citet{kryscinski2019evaluating} finetune a BERT model on heuristically-created data with six types of rule-based text transformations for factual consistency checking. \citet{wang2020asking} propose a framework for measuring inconsistencies in abstractive summarization by answering questions based on both generated summaries and documents.

\section{Proposed Approach}
In this section, we describe our procedure of introducing artificial errors in the datasets for training and propose our end-to-end error corrector model. 
\subsection{Dataset of Artificial Corruptions}
\label{sec:dataset}

Inspired by a recent study of error types made by state-of-the-art summarization system, we artificially created a weakly-supervised training dataset based on the text transformations proposed by \citet{kryscinski2019evaluating}. 

Given a source text $d$ and the reference summary $s$, we corrupt the reference summary into an inconsistent summary $s'$ with a randomly sampled corruption rule (described below) with probability $\alpha$; otherwise, we keep $s'=s$ with probability $1-\alpha$. We set $\alpha=0.3$ to match the factuality error rate in real abstract summaries based on a recent study \citep{cao2018faithful}. The training data consists of triplets $(s', s, d)$. 

\definecolor{MyGreen}{rgb}{0.76,0.88,0.71}
\definecolor{MyBlue}{rgb}{0.18,0.33,0.80}
\definecolor{MyOrange}{rgb}{0.97,0.80,0.68}

\begin{table}[t]
\small
\renewcommand{\arraystretch}{1.2}
\setlength\tabcolsep{2.5pt}
\centering
\begin{tabular}{|p{7.3cm}|}
  \hline
  {\bf Source}: \\
  (CNN) Gastrointestinal illness has gripped 100 people on the cruise ship Celebrity Infinity, according to a report from the Centers for Disease Control. Of the ship's 2,117 passengers, 95 have suffered from vomiting, diarrhea and other symptoms, the CDC said. (...) \\
  \hline
  {\bf Reference Summary}: \\
  \textcolor{MyBlue}{\bf 100} passengers and crew members have been sickened on Celebrity Infinity. The ship, which is based on the West Coast, left San Diego in late March . 
  \\
  {\bf Corrupted Summary}: \\
  \textcolor{MyBlue}{\bf 95} passengers and crew members have been sickened on Celebrity Infinity. The ship, which is based on the West Coast, left San Diego in late March . 
  \\
  \hline
\end{tabular}
\caption{\label{table:train_example} An example of a \textit{Number} corruption in the training set. The incorrect number ``95'' also appears in the source document.}
\end{table}

\paragraph{Error Corruptions} \label{sub-section:error_corruption}
Four types of errors are used to create the inconsistent summaries: \textit{Entity}, \textit{Number}, \textit{Date}, and \textit{Pronoun} errors. They are the most common types of errors in abstractive summaries based on our manual inspection of 100 abstractive system-generated summaries that are sampled from the dataset of \citet{kryscinski2019evaluating} (henceforth, the \textsc{K2019} dataset). Unlike \citet{kryscinski2019evaluating}, we corrupt the reference summary rather than sentences sampled from the source document. 

In the first four types of error constructions, we utilize a swapping strategy to introduce errors. For \textit{Entity}, \textit{Number}, and \textit{Date} swapping, one entity in the reference summary is selected and swapped with another random entity of the same type\footnote{All the entities are extracted using a pre-trained NER model in spaCy \url{https://spacy.io/}. } in the source document. 
For \textit{Pronoun} swapping, one pronoun was extracted and swapped with another one of a matching syntactic case. Table~\ref{table:train_example} shows one example of a corruption.



\subsection{Training Objective and Models}
\label{sec:objective}
With the artificial training data consisting of triplets $(s', s, d)$, the goal of the corrector is to generate the correct summary $s$ based on the inconsistent summary $s'$ and the source $d$. This can be expressed as a problem of maximizing the likelihood of $P(s|s',d)$ in an encoder-decoder model. We concatenate $s'$ and $d$ as input to the encoder ($s'$ and $d$ are separated by a separation token) and train the decoder to generate $s$. 

We use BART \cite{lewis2019bart} as the basis of our summary corrector because of its demonstrated level of performance on conditional text generation tasks. BART is a sequence-to-sequence auto-regressive transformer model that is pre-trained as a denoising auto-encoder. One appealing aspect about BART is that it is pre-trained on a denoising task. Specifically, given an input sentence that is corrupted by text infilling, token deletion as well as other text transformations, BART is trained to output the original sentence. This pre-training task is similar to our summary correction task in which we can regard the corrupted or generated summary as the noisy input and in this case the noise is the inconsistent content in the summary.

\section{Experiments}
\subsection{Evaluation Tasks and Measures}
\label{sec:eval_metrics}
We evaluate our model on two tasks: factual consistency checking and error correction.

\paragraph{Factual consistency checking} For this task, the model needs to classify each original input summary as \emph{consistent} or \emph{inconsistent} with respect to the source text. It is thus a binary classification task for which we report \textbf{accuracy}, as well as \textbf{precision}, \textbf{recall}, and \textbf{F1}.

We interpret the output of our corrector model as a classification decision as follows. If the corrector makes any change to the original input summary, we consider this to be a prediction of the \emph{inconsistent} class. Otherwise, the corrector makes no change and we consider this a prediction of the \emph{consistent} class.

\begin{table}[t!]
\renewcommand{\arraystretch}{1.1}
\begin{center}
\begin{tabular}{c|c|ccc}
\toprule
\multirow{2}{*}{~} & \bf \multirow{2}{*}{\makecell{Overall \\ Acc.}} & \multicolumn{3}{c}{\bf Consistency checking} \\ 
~ & \bf ~ & Prec. & Recall & F1 \\ 
\midrule
Corrupted & \multirow{2}{*}{84.38\%} & 0.79 & 0.95 & 0.86 \\

Clean & ~ & 0.93 & 0.74 & 0.82 \\
\bottomrule
\end{tabular}
\end{center}
\caption{\label{table:analysis1} Performance of our model on consistency checking on our test set of artificial corruptions. Corrupted and clean refer to the subsets of the test set that were artificially perturbed or not perturbed, respectively. 
}
\end{table}

\paragraph{Error correction} For this task, the model must correct inconsistencies in the original summary (in any) with respect to the source text.

We define \textbf{correction accuracy} as the proportion of original summaries that are correctly changed by our corrector. On our artificial test set, an input summary is considered successfully corrected if the corrected summary matches the reference summary exactly.
For the \textsc{K2019} dataset, no reference corrections are available. We instead conducted a human evaluation to check the consistency of the corrected output. We read the original and corrected summaries as well as the source document to determine whether a summary is successfully corrected by our model.

\subsection{Datasets}
We use two datasets for our experiments. The first is the dataset of artificial corruptions described in Section~\ref{sec:dataset}, which we create by taking samples from the CNN/DailyMail dataset. There are in total 287,227 samples in the training set, and we corrupted 30\% of them (85,583). This results in 16,858/35,113/13,408/20,204 date/entity/number/pronoun corrupted samples respectively. We refer the other 201,644 training samples as clean samples. We also create artificial validation and test set for model selection and evaluation. In the test set, there are 5,780 corrupted samples and 5,710 clean samples.

The second dataset we use is the \textsc{K2019} test set of \citet{kryscinski2019evaluating}.
This dataset contains 503 summaries generated by different recent neural abstractive summarizers, which have been manually labeled for whether they contain an inconsistency.

We evaluate our model on both datasets. We did not use baselines for the artificial test set since it is simply used as a check to demonstrate our model's performance in the artificial setting. The more meaningful evaluations are on K2019 consistency checking and error correction.

\subsection{Corrector Training Details}
We use the BART implementation from fairseq as the basis of our corrector.\footnote{\url{https://github.com/pytorch/fairseq/blob/master/examples/bart}}
The pre-trained BART model is fine-tuned on our training dataset for 10 epochs as described in Section \ref{sec:objective}. The learning rate is set to 3e-5. All our experiments is done on 4 NVIDIA Tesla V100 GPUs. The training process takes about 12 hours.

\begin{table}[t!]
\renewcommand{\arraystretch}{1.1}
\begin{center}
\begin{tabular}{c|c|c}
\toprule
Model & \thead{\bf Accuracy \\ (\textit{weighted})} & \bf F1-score \\ 
\midrule
BERT+MNLI & 51.39\% & 0.86 \\ 
BERT+FEVER & 52.07\% & 0.88 \\ 
FactCC & \bf 72.65\%  & \bf 0.86 \\ 
FactCCX & 72.88\% & 0.87 \\ 
\midrule
Our model & 66.46\% &0.83 \\ 
\bottomrule
\end{tabular}
\end{center}
\caption{\label{table:analysis2} Factual consistency checking performance on the \textsc{K2019} test set. The F1-score reported from our model is the micro-average F1-score.}
\end{table}

\begin{table*}[t]
\small
\renewcommand{\arraystretch}{1.2}
\setlength{\belowcaptionskip}{-0.40cm}
\centering
\begin{tabular}{p{15.5cm}}
  \toprule
  \emph{Article}: Jerusalem (CNN)The flame of remembrance burns in Jerusalem, and a song of memory haunts Valerie Braham as it never has before. (...) ``Now I truly understand everyone who has lost a loved one,'' Braham said. \textcolor{MyBlue}{Her husband, Philippe Braham, was one of 17 people killed in January's terror attacks in Paris.} He was in a kosher supermarket when a gunman stormed in, killing four people, all of them Jewish. (...) \\ 
  \emph{Original}: {\bf Valerie} braham was one of 17 people killed in january's terror attacks in paris. \emph{(inconsistent)} \\ 
  \emph{Corrected}: {\bf Philippe} braham was one of 17 people killed in january's terror attacks in paris. \emph{(consistent)} \\ 
  \hline
  \emph{Article}: (...) Thursday's attack by al-Shabaab militants killed 147 people, including 142 students, three security officers and two university security personnel. \textcolor{MyBlue}{The attack left 104 people injured, including 19 who are in critical condition, Nkaissery said.} (...) \\ 
  \emph{Original}: {\bf 147} people, including 142 students, are in critical condition. \emph{(inconsistent)} \\ 
  \emph{Corrected}: {\bf 19} people, including 142 students, are in critical condition. \emph{(inconsistent)} \\ 
  \hline
  \emph{Article}: (CNN) Officer \textcolor{MyBlue}{Michael Slager}'s five-year career with the North Charleston Police Department in South Carolina ended after he resorted to deadly force following a routine traffic stop. (...) His back is to Slager, who, from a few yards away, raises his gun and fires. \textcolor{MyBlue}{Slager is now charged with murder.} The FBI is involved in the investigation of the slaying of the father of four. (...) \\ 
  \emph{Original}: {\bf Slager} is now charged with murder. \emph{(consistent)} \\ 
  \emph{Corrected}: {\bf Michael Slager} is now charged with murder. \emph{(consistent)} \\
  \hline
  \emph{Article}: \textcolor{MyBlue}{(CNN)The announcement this year of a new, original Dr. Seuss book sent a wave of nostalgic giddiness across Twitter,} and months before publication, the number of pre-orders for ``What Pet Should I Get?'' continues to climb. (...) \textcolor{MyBlue}{It features the spirited siblings from the beloved classic ``One Fish Two Fish Red Fish Blue Fish'' and is believed to have been written between 1958 and 1962.} (...) \\
  \emph{Original}: {\bf Seuss} book sent a wave of nostalgic giddiness across twitter. \emph{(consistent)} \\ 
  \emph{Corrected}: {\bf ``One Fish Two Fish Red Fish Blue Fish''} book sent a wave of nostalgic giddiness across twitter. \emph{(inconsistent)} \\
  \bottomrule
\end{tabular}
\caption{\label{table:samples} Examples of applying our corrector to the output of a summarizer. In the first example, the original inconsistent summary is successfully corrected. In the second example, the summary remains false after correction. The third and fourth examples show changes made on consistent summaries. Colored content in articles are support for summaries.
}
\end{table*}

\begin{table}[t!]
\renewcommand{\arraystretch}{1.15}
\setlength{\belowcaptionskip}{-0.46cm}
\begin{center}
\begin{tabular}{c|c|cc}
\toprule
\multirow{2}{*}{\bf Input} & \bf \multirow{2}{*}{\# Samples} & \multicolumn{2}{c}{\bf After Correction} \\ 
~ & \bf ~ & \emph{ cons.} & \emph{incons.} \\ 
\midrule
\emph{consistent} & 441 & 436 & 5 \\
\midrule
\emph{inconsistent} & 62 & 11 & 51 \\
\bottomrule
\end{tabular}
\end{center}
\caption{\label{tab:human-eval} Human evaluation results on error correction on the \textsc{K2019} dataset.}
\end{table}

\section{Results}
\label{sec:results}
\paragraph{Artificial corruptions}
Table~\ref{table:analysis1} shows the consistency checking performance of our corrector model on our artificial test set. The high classification accuracy and F1 scores indicate that our model is able to identify these artificially injected errors.

For error correction, among the 5780 corrupted summaries in the test set, 62.13\% are corrected by the model to exactly match the reference summary. For the 5710 clean summaries, the model made changes to 26.27\% of them, which results in 73.73\% correction accuracy on clean summaries. These results show that the model is able to correct majority of the test samples even under our strict evaluation measure.


\paragraph{\textsc{K2019}}
Table~\ref{table:analysis2} shows the  consistency checking results on the \textsc{K2019} test set. Our model is better than the BERT model and slightly worse compared with the FactCC model.


As for correction performance, Table~\ref{tab:human-eval} shows the evaluation result of our human evaluation. Among 62 inconsistent summaries in the test set, the corrector model made changes to 19 summaries, of which 11 were successfully corrected and 7 remained inconsistent.
For the remaining 441 consistent summaries in the test set, changes are made to 39 summaries and the model changed the meaning of 5 samples. In conclusion, with 17.74\% probability that our model can successfully correct an inconsistent summary and 1.13\% probability that it will corrupt a consistent one. Compared with the correction rate of 62.13\% on the artificial test set, much lower correction rate on the real test set  suggests that there is still a gap between the two settings. The error types in the training set are not able to represent the diverse errors made by summarization systems.

\paragraph{Output Analysis} Table~\ref{table:samples} shows several input and output summaries of our corrector model together with the source document fragments. In the second example, the model correctly replaced 147 with 19, but was not able to correctly remove ``including 142 students'', which is a larger modification to the original summary. More examples can be found in the Appendix.

\section{Conclusions}

In this paper, we proposed a novel approach to correct inconsistent content in summaries generated by abstractive summarization models. We train an end-to-end correction model with artificial examples created by corrupting reference summaries. Our model achieved promising performance on our artificial test set and outperformed previous models on the manually annotated test set by wide margins. Our human evaluation indicates that our model is able to correct some factually inconsistent summaries generated by abstractive summarization model. However, low recall on the inconsistent summaries and false positive samples remain as challenges.

\section*{Acknowledgments}
This research was supported by the Canada CIFAR AI Chair program, the Natural Sciences and Engineering Research Council of Canada (NSERC) and the Fonds de recherche du Qu\'{e}bec -- Nature et technologies (FRQNT). 
We would also like to thank Compute Canada for providing us computing resources.

\bibliographystyle{acl_natbib}
\bibliography{emnlp2020}

\appendix
\newpage

\section{Appendix}
\subsection{Summary Correction Examples}
Table \ref{table:samples} shows examples of generated summaries and outputs from our corrector. We put more examples here:

\begin{enumerate}
	\item \emph{Article}: (CNN)In case you needed a reminder that President Barack Obama isn\'t running for office again, he just alienated not only Republicans, who have largely resented him from day one, but the progressive base of Democratic voters. Obama has argued with the progressive potentate Elizabeth Warren, calling her ``wrong'' on trade policy. (...) \\
    \emph{Original}: {\bf Obama } has argued with the progressive potentate elizabeth warren. \emph{(consistent)} \\ 
    \emph{Corrected}: {\bf President Obama } has argued with the progressive potentate elizabeth warren. \emph{(consistent)} 
	\item \emph{Article}: (The Hollywood Reporter)The author of a 2006 novel has accused the ``Avengers'' director and ``Cabin'' director Drew Goddard of stealing his idea. (...) Gallagher is basing his claim on the works' similar premises: Both feature a group of young people terrorized by monsters while staying at a cabin in what is revealed to be (spoiler alert) a horror-film scenario designed by mysterious operators. (...) \\
    \emph{Original}: {\bf Gallagher } is basing his claim on the works' names and personalities. \emph{(inconsistent)} \\ 
    \emph{Corrected}: {\bf Peter Gallagher } is basing his claim on the works' names and personalities. \emph{(inconsistent)}
	\item \emph{Article}: (CNN)Too little, too late. (...) After the story of the statue caught fire online this weekend, Poulin publicly apologized Monday for his ``most unsettling sculpture'' in a letter to The Hollywood Reporter. (...)\\
    \emph{Original}: {\bf Poulin} publicly apologized for his ``most unsettling sculpture'' in a letter to the hollywood reporter. \emph{(consistent)} \\ 
    \emph{Corrected}: {\bf Dave Poulin} publicly apologized for his ``most unsettling sculpture'' in a letter to the hollywood reporter. \emph{(consistent)} 
	\item \emph{Article}: (CNN)If I had to describe the U.S.-Iranian relationship in one word it would be ``overmatched''. (...) America is alienating some of our closest allies because of the Iran deal, and Iran is picking up new ones and bolstering relations with old ones who are growing more dependent because they see Iran's power rising. (...) \\
    \emph{Original}: {\bf Iran} is alienating some of our closest allies because of the iran deal, and iran is picking up new ones. \emph{(inconsistent)} \\ 
    \emph{Corrected}: {\bf America} is alienating some of our closest allies because of the Iran deal, and Iran is picking up new ones. \emph{(consistent)} 
	\item \emph{Article}: (...) McHenry quickly issued an apology, blaming the incident on a moment of intense frustration but admitting her mistake and accepting responsibility. (...) \\
    \emph{Original}: {\bf Mchenry} apologizes to the incident on a moment of intense frustration. \emph{(consistent)} \\ 
    \emph{Corrected}: {\bf Britt Mchenry} apologizes to the incident on a moment of intense frustration. \emph{(consistent)} 
	\item \emph{Article}: Boston (CNN)When the bomb went off, Steve Woolfenden thought he was still standing. That was because, as he lay on the ground, he was still holding the handles of his son's stroller. He pulled back the stroller's cover and saw that his son, Leo, 3, was conscious but bleeding from the left side of his head. (...) \\
    \emph{Original}: {\bf Steve woolfenden}, 3, was conscious but bleeding from the left side of his head. \emph{(inconsistent)} \\ 
    \emph{Corrected}: {\bf Leo Woolfenden}, 3, was conscious but bleeding from the left side of his head. \emph{(consistent)} 
	\item \emph{Article}: (CNN)Mercedes driver and F1 championship leader Lewis Hamilton stole pole position for Sunday's Chinese Grand Prix from teammate and fierce rival Nico Rosberg in dramatic fashion. (...) He did, however, find time to congratulate fellow German driver Sebastian Vettel, who will start in third after the Ferrari driver surprisingly won the Malaysian GP two weeks ago. (...) \\
    \emph{Original}: {\bf Sebastian vettel} won the malaysian gp two weeks ago. \emph{(consistent)} \\ 
    \emph{Corrected}: {\bf Nico Rosberg} won the malaysian gp two weeks ago. \emph{(inconsistent)} 
	\item \emph{Article}: (CNN)At least 21 people were killed during a shipwreck off the northern coast of Haiti, the country's civil protection directorate told CNN on Thursday. (...) So far, 11 victims -- eight men and three women -- have been identified, Celestin said. (...) \\
    \emph{Original}: The 21 people are {\bf three} women and three women have been identified. \emph{(inconsistent)} The 21 people are {\bf eight} women and three women have been identified. \\ 
    \emph{Corrected}: \emph{(inconsistent)} 
	\item \emph{Article}: (CNN)Oklahoma Gov. Mary Fallin signed a bill on Friday that would allow the state to perform executions with nitrogen gas if lethal injection is ruled unconstitutional or becomes unavailable. Nitrogen causes a quick loss of consciousness and then death from lack of oxygen, Fallin's office said in a press release. (...) \\
    \emph{Original}: {\bf Nitrogen} causes a quick loss of consciousness and then death from lack of oxygen. \emph{(consistent)} \\ 
    \emph{Corrected}: {\bf Netherlands} causes a quick loss of consciousness and then death from lack of oxygen. \emph{(inconsistent)} 

\end{enumerate}

\end{document}